\UseRawInputEncoding
\documentclass[10pt,twocolumn,letterpaper]{article}

\usepackage{iccv}
\usepackage{times}
\usepackage{epsfig}
\usepackage{graphicx}
\usepackage{amsmath}
\usepackage{amssymb}

\usepackage{graphicx}
\usepackage{booktabs}

\usepackage{algorithm}
\usepackage{algorithmic}
\usepackage{blindtext}
\usepackage{caption}
\usepackage{float}
\usepackage{multirow}
\usepackage{url}
\usepackage{makecell}
\usepackage{multicol}
\usepackage{color}
\usepackage[switch]{lineno}

\usepackage[pagebackref=true,breaklinks=true,letterpaper=true,colorlinks,bookmarks=false]{hyperref}

\iccvfinalcopy 

\def\iccvPaperID{9144} 

\ificcvfinal\fi

\begin{document}

\title{Creative Birds: Self-Supervised Single-View 3D Style Transfer}


\author{{Renke Wang$^{1*}$\, Guimin Que$^{1*}$, Shuo Chen$^{2\dagger}$, Xiang Li$^3$, Jun Li$^{1\dagger}$, Jian Yang$^1$}\\
\\
{$^1$PCA Lab, Nanjing University of Science and Technology, China}\\
{$^2$RIKEN, $^3$Nankai University}\\
{\tt\small \{wrk226,qgm226131,xiang.li.implus,junli,csjyang\}@njust.edu.cn 
   shuo.chen.ya@riken.jp}\\
{\tt\small $^{*}$Contributes equally\ \ \ \ \ \ $^{\dagger}$Corresponding author}\\
}

\ificcvfinal\thispagestyle{empty}\fi

\twocolumn[{
    \renewcommand\twocolumn[1][]{#1}
    \maketitle
    \begin{center}
     \vskip -0.2in
        \captionsetup{type=figure}
        \includegraphics[width=1\textwidth]{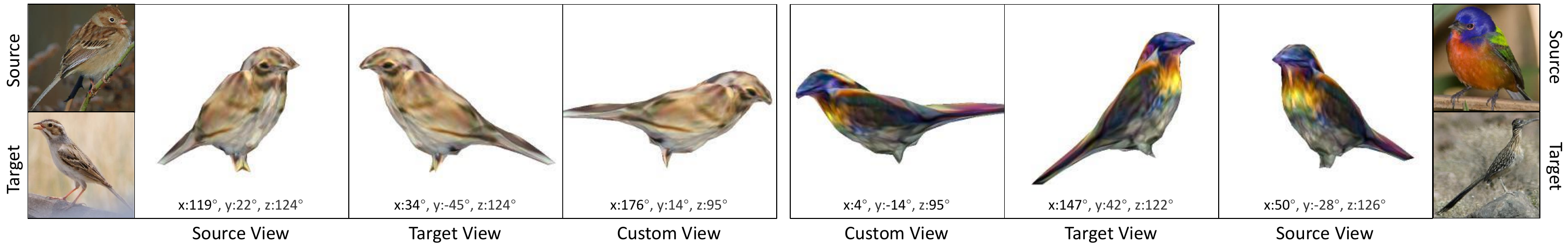}
        \vskip -0.1in
        \captionof{figure}{We propose a method to generate 3D birds by transferring both the geometric and textural style of target image to source. The left and right pictures show the 3D birds between populations of the same and different species, respectively.}
        \vskip 0in
   \label{fig:Fig1}
    \end{center}
}]

\begin{abstract}
In this paper, we propose a novel method for single-view 3D style transfer that generates a unique 3D object with both shape and texture transfer. 
Our focus lies primarily on birds, a popular subject in 3D reconstruction, for which no existing single-view 3D transfer methods have been developed.
The method we propose seeks to generate a 3D mesh shape and texture of a bird from two single-view images. To achieve this, we introduce a novel shape transfer generator that comprises a dual residual gated network (DRGNet), and a multi-layer perceptron (MLP). DRGNet extracts the features of source and target images using a shared coordinate gate unit, while the MLP generates spatial coordinates for building a 3D mesh. We also introduce a semantic UV texture transfer module that implements textural style transfer using semantic UV segmentation, which ensures consistency in the semantic meaning of the transferred regions. This module can be widely adapted to many existing approaches. Finally, our method constructs a novel 3D bird using a differentiable renderer. Experimental results on the CUB dataset verify that our method achieves state-of-the-art performance on the single-view 3D style transfer task. Code is available in \href{https://github.com/wrk226/creative_birds}{here}.

\end{abstract}

\vspace{-4mm} 

\section{Introduction}
\label{sec:intro}
Neural image style transfer has received increasing attention in computer vision communities due to its remarkable success in many automatic creations. These applications mainly belong to a 2D style transfer, which transfers the artistic style of one reference image to another content image. Recently, it is extended to transfer the shape and texture style of one 3D object to another for editing 3D content in augmented reality and virtual reality \cite{yin2021_3DStyleNet}. However, this extension needs to acquire the 3D information on objects which suffers from the significant problem of being arduous, high-cost, and time-consuming. In contrast, taking a single-view object image is easier and cheaper than 3D data as cameras (\eg, mobile phones) are used widely in everyday life. Therefore, we address a new task of \emph{generating a 3D object with novel shape and texture by transferring one single-view object image to another}, instead of acquiring 3D input information. We refer to this task as a single-view 3D style transfer.
Prior works have primarily handled either 3D style transfer or single-view 3D reconstruction but have not simultaneously considered both for the aforementioned task. The former heavily depends on the 3D information of the object as an input, which is often expensive and time-consuming. For example, 3DStyleNet \cite{yin2021_3DStyleNet} requires training on a large dataset of 3D shapes, while Neural Renderer \cite{kato2018neural} transfers the style of an image onto a 3D mesh. In addition, 3D scene stylization \cite{huang2021newview,mu2022newview} generates stylized novel view synthesis from multiple RGB images given an arbitrary style. 
The latter approach can easily implement 3D object reconstruction using single-view images. Classical methods predict the reconstructed information of a 3D object from the single-view images \cite{cmrKanazawa2018,Henderson2020SI3DR,kato2019learning,Wu2020image23Dobject,Hu_2021_3Dmesh}. Specifically, category-Specific mesh reconstruction (CMR) \cite{cmrKanazawa2018} learns to recover the 3D mesh shape, texture, and camera pose of an object from the single images of one object category. Following CMR, unsupervised mesh reconstruction (UMR) \cite{li2020self} uses a self-supervision with semantic part consistency to predict a 3D object. As UMR does not require 3D information, a union between the two approaches becomes a viable strategy to tackle the single-view 3D style transfer task.

\begin{figure*}[t]
\centering
\includegraphics[width=0.83\linewidth]{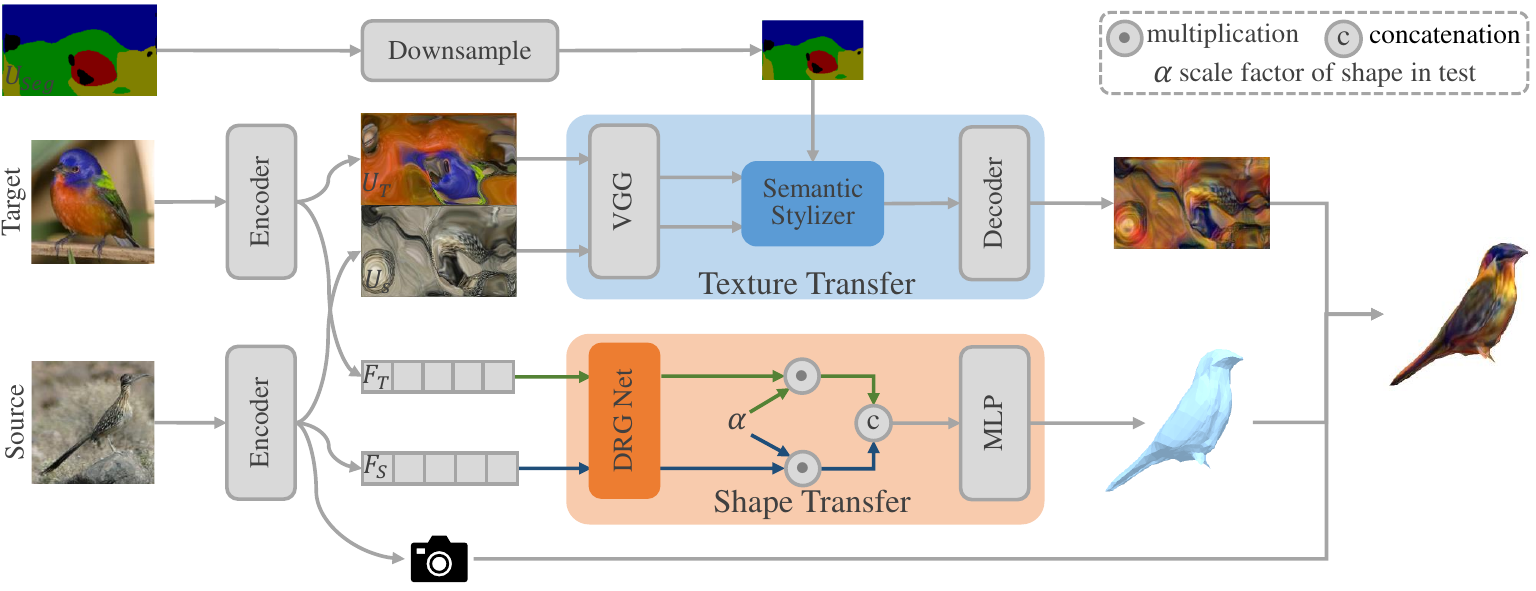}
\vskip -0.1in
\caption{The proposed single-view 3D style transfer pipeline employs a shared encoder from UMR \cite{li2020self} and a semantic UV segmentation mask $U_{seg}$ from SCOPS \cite{Hung2019scops}, as well as a semantic texture transfer and shape transfer generator. The DRGNet and Semantic Stylizer are shown in Figs. \ref{fig:drgnet} and \ref{fig:semtransfer}, respectively.}\label{fig:pipeline}
\vskip -0.15in
\end{figure*}

In this paper, we primarily focus on the novel 3D bird generation as it is a popular example \cite{cmrKanazawa2018,Badger20203Dbirds,Wang2021image2shape,li2020self} in single-view 3D reconstruction, especially UMR \cite{li2020self}. Specifically, given a pair of single-view source and target bird images within and between species, we explore plausible 3D bird creation by transferring the geometric and textural styles of the target bird to the source, as shown in Fig. \ref{fig:pipeline}. First, we employ the reconstruction network in UMR \cite{li2020self} as an encoder to output the shape features, UV textures, and camera poses of the target and source bird images, respectively. Second, we propose a shape transfer generator for implementing the geometric transfer which consists of a dual residual gated network (DRGNet), a scale factor, and a multi-layer perceptron (MLP). DRGNet is designed with a shared coordinate gate unit to select both the target and source shape features, separately cascading single-layer networks and incorporating residual connections into their networks in Fig. \ref{fig:drgnet}. The scale factor controls the degree of the transformation from the source to the target. MLP generates the spatial coordinates necessary for building a 3D mesh. Third, we introduce a semantic UV texture transfer that exploits switched gates to respectively control whether the part textural transfer or not using AdaIN \cite{Huang2017adain}, LST \cite{Li2019lineartransfer} and EFDM \cite{zhang2022efdm}. Finally, a novel 3D bird is constructed by using a differentiable renderer \cite{liu2019soft} with the pose of the source, the mesh, and the semantic stylized UV map. Overall, our contributions are summarized as follows:
\begin{itemize}
\item We propose a shape transfer generator consisting of a dual residual gated network, a scale factor, and a multi-layer perceptron for the geometric style transfer. 
\item We present a semantic UV transfer method to utilize switched gates to respectively control whether the part textural transfer or not by using AdaIN \cite{Huang2017adain}, LST \cite{Li2019lineartransfer}, and EFDM \cite{zhang2022efdm}, and employing semantic UV segmentation for the texture style transfer.
\item To the best of our knowledge, we are the first to address the single-view 3D style transfer task by combining the shape and texture transfer to automatically create novel 3D birds from images in Fig. \ref{fig:Fig1}.
\end{itemize}

\section{Related Work}
Here, we mainly review some related tasks about the style transfer from single-view images to 3D objects, including single-view 3D reconstruction \& deformation, and 2D \& 3D style transfer.



\textbf{Single-view 3D reconstruction \& deformation.}
Single-view 3D reconstruction aims to generate a 3D model from a single 2D image. Early works \cite{wang2018pixel2mesh, pan2019deep} trained the model using the image and ground truth mesh pairs, which are usually limited and expensive to obtain. Based on the differentiable renderers \cite{kato2018neural,liu2019soft,ravi2020accelerating}, weakly supervised methods become possible \cite{Wu2020image23Dobject,zhang2021learning,sun2022neural}, but typically require multi-view images or known camera parameters. CMR \cite{cmrKanazawa2018} avoids these requirements by constructing inter-class template mesh, 
but still needs to be initialized by the landmark. UMR \cite{li2020self} avoids the need for landmarks by maintaining semantic correspondence between objects of the same class. Using UMR, our model gains the capability to perform the 3D reconstruction.


The goal of 3D deformation is to deform the shape of the target object to the source. Some early works tried to set controllable points \cite{sumner2004deformation, sorkine2007rigid} or cage \cite{calderon2017bounding} to deform the shape of the object with manual control. 
To avoid human intervention, some works tried to find the correspondence between source and target through building deformable cage \cite{yifan2020neural},  encoding deformation process \cite{Sung2020DeformSyncNet} and detecting 3D keypoints \cite{jakab2021keypointdeformer}. We will compare our shape transfer module with some of these works using the 3D reconstructed shape.



\textbf{2D \& 3D style transfer.}
2D style transfer has been extensively explored with many proposed methods such asAdaIN \cite{Huang2017adain}, LST \cite{Li2019lineartransfer}, Adaattn\cite{liu2021adaattn}, InST \cite{jcyang2022InST}, and EFDM \cite{zhang2022exact}. These methods can be easily extended to UV texture maps as they are also 2D images.
Following this strategy, we explore a semantic UV texture transfer method for the diversity of the stylized UV texture.

For 3D style transfer, early works \cite{kato2018neural,liu2018paparazzi,mordvintsev2018dip} reconstructed 3D meshes from single-view or multi-view images and used differential rendering methods to transfer a style image into the 3D mesh. 3DStyleNet \cite{yin2021_3DStyleNet} focuses on shape transformation and still uses the texture of the source in the transferred mesh. Other methods perform 2D-like style transformation for point clouds \cite{huang2021newview,mu2022newview} or implicit fields \cite{huang2022stylizednerf,chiang2022stylizing,zhang2022arf} without considering the style of the 3D shape.

In contrast, we study both 3D shape and UV texture transfer from the target image to a source image for creating a 3D object. The work most similar to ours is 3D portrait stylization \cite{han2021exemplarbased}, which produces a stylized 3D face composed of geometric and texture style transfer. The geometric transfer reconstructs a 3D geometry from the source face image, utilizing facial landmarks translation between the source and target face images to guide the deformation of the dense 3D face geometry. The texture transfer performs style transfer on the target texture by using a differentiable renderer in a multi-view manner. In contrast to this work, we design a shape transfer network to directly generate a novel 3D mesh and introduce a semantic UV texture transfer method to obtain a new UV texture. More importantly, our work can handle the single-view 3D style transfer task for novel 3D object generation.



\vspace{-1mm}
\section{Single-view 3D Style Transfer}
In this section, we aim to extract 2D information from two bird images and fuse these features to construct a novel 3D bird. To achieve this, we propose a single-view 3D style transfer framework that automatically generates a plausible 3D bird, as illustrated in Fig. \ref{fig:pipeline}. 
In our method, we utilize the pre-trained ResNet-18 from the UMR paper \cite{li2020self} as a feature extractor to extract shape features. We also adopt the texture predictor  and camera predictor from UMR to predict the corresponding textures and camera parameters. In the following sections, we will refer to these modules as the encoder in our network.
Given a source bird image $S\in \mathbb{R}^{256\times256\times3}$ and a target bird image $T\in \mathbb{R}^{256\times256\times3}$, the encoder outputs the camera pose $C_S\in \mathbb{R}^{7}$, $C_T\in \mathbb{R}^{7}$, shape features $F_S\in \mathbb{R}^{512}$, $F_T\in \mathbb{R}^{512}$ and the UV texture flows $U_S\in \mathbb{R}^{128\times256\times3}$, $U_T\in \mathbb{R}^{128\times256\times3}$, respectively. 
Concurrently, we construct a semantic UV segmentation $U_{\text{seg}}\in \{1,2,3,4,5\}^{128\times256}$ from UMR \cite{li2020self} using SCOPS \cite{Hung2019scops}. Following the encoder and the UV segmentation, our framework mainly consists of a shape transfer generator from $F_S$ to $F_T$ and a semantic texture transfer from $U_S$ to $U_T$.


\subsection{Shape Transfer from Images to 3D Mesh}
\label{sec:shapetransfer}
Here, we aim to generate a novel 3D shape from source and target bird images $S$ and $T$. Using the previously described encoder, we obtain their shape features $F_S$ and $F_T$. Specifically, we explore a shape transfer to generate a 3D mesh by progressively adjusting and filtering the shape features $F_S$ and $F_T$. It includes a dual residual gated network (DRGNet), a scale factor, and a multi-layer perceptron.  
\begin{figure}[t]
\centering
\includegraphics[width=0.95\linewidth]{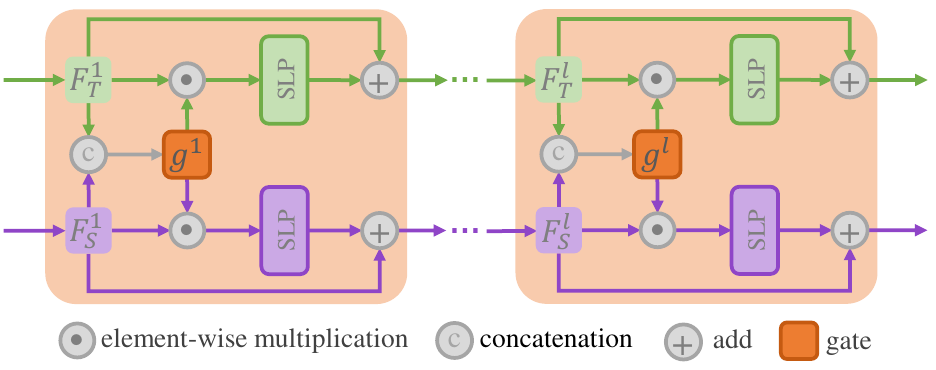}
\vskip -0.1in
\caption{The Dual Residual Gated Network consists of stacked dual residual gate units that refine the input features $F_S$ and $F_T$, and outputs the refined features $F_S^L$ and $F_T^L$.}\label{fig:drgnet}
\vskip -0.15in
\end{figure}

\textbf{Dual Residual Gated Network} is designed by stacking dual residual gate units as shown in Fig. \ref{fig:drgnet}. Following the shape features $F_S$ and $F_T$, we build dual source and target branches. Since these features are extracted from the same encoder, their coordinates have potential correspondences. So we design a gate shared in the dual branches to select the features $F_S$ and $F_T$ in the same coordinates for benefiting the shape transfer. Sequentially, we connect two single-layer perceptrons in the dual branches, respectively. Simultaneously, a residual connection \cite{he2016deep} is added to alleviate over-fitting, and gradient vanishing, and prevent distortion caused by the unselected features. Beginning with the initial inputs $F_S^1=F_S$ and $F_T^1=F_T$, the $l$-th unit of dual residual gate is described as:
\begin{align}
g^l&=\text{Sigmoid}\left(W^l\cdot [F_S^l,F_T^l]\right), \nonumber\\
F_S^{l+1}&=F_S^l+\text{ReLU}\left(\text{BatchNorm}\left(W_{S}^l\cdot [F_S^l\odot g^l]\right)\right), \label{eq:drgu} \\
F_T^{l+1}&=F_T^l+\text{ReLU}\left(\text{BatchNorm}\left(W_{T}^l\cdot [F_T^l\odot g^l]\right)\right), \nonumber
\end{align}
where $W^l\in \mathbb{R}^{512\times1024}$, $W_{S}^l\in \mathbb{R}^{512\times512}$ and $W_{T}^l\in \mathbb{R}^{512\times512}$ are the weights of the $l$-th unit, and Sigmoid, Tanh, and ReLU are the activation functions, respectively. To enhance the representation of specific individual traits, we stack the gated unit in Eq. \eqref{eq:drgu} $L$ times in our DRGNet for refining the features progressively, and it outputs the features $F_S^L\in \mathbb{R}^{512}$ and $F_T^L\in \mathbb{R}^{512}$. In this paper, the stacked number is set to $L=8$.

\textbf{Scale Factor} $\alpha$ is to control the proportion of the features $F_S^L$ and $F_T^L$ during testing, while it is set to one during training. Using the scale factor, the features are computed by
\begin{align}
[\overline{F}_S, \overline{F}_T]=[(1-\alpha)\cdot F_S^L, (1+\alpha)\cdot F_T^L] . 
\label{eq:scale}
\end{align}
where $-1\leq\alpha\leq 1$, $\overline{F}_S\in \mathbb{R}^{512}$ and $\overline{F}_T\in \mathbb{R}^{512}$.

\textbf{Multi-layer Perceptron} is employed to fuse the features $\overline{F}_S$ and $\overline{F}_T$. It is a simple two-layer neural network with the ReLU activation function, which is defined by
\begin{align}
O=W_o\cdot\text{ReLU}\left(W_f^2\cdot\text{ReLU}\left(W_f^1\cdot [\overline{F}_S, \overline{F}_T]\right)\right),
\label{eq:mlp}
\end{align}
where $W_f^1\in \mathbb{R}^{512\times1024}$, $W_f^2\in \mathbb{R}^{512\times512}$, and $W_o\in \mathbb{R}^{3\times372\times512}$ are the network parameters, and $O\in \mathbb{R}^{3\times372}$ is the output of the multi-layer perceptron.

Since most birds are symmetric, we only predict the left half of the object in the output $O$, with a total of 372 vertices. We then obtain an additional 270 coordinates of the symmetrical vertices, as the remaining 102 vertices are on the symmetrical plane. Finally, we use all of the 642 three-dimensional coordinates to build the 3D mesh by employing a differentiable renderer, \ie, soft rasterizer \cite{liu2019soft}.

\subsection{Semantic UV Texture Transfer}
\label{sec:texturetransfer}
\begin{figure}[t]
\centering
\includegraphics[width=0.87\linewidth]{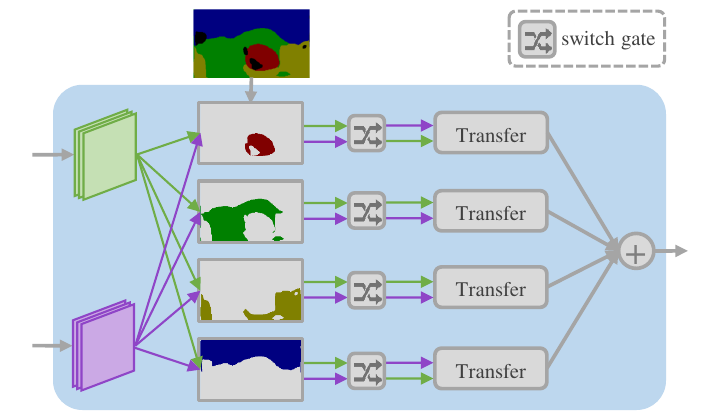}
\vskip -0.05in
\caption{The Semantic Stylizer uses the semantic UV mask produced by SCOPS \cite{Hung2019scops} to maintain consistent semantic meaning of the transferred regions during style transfer.}\label{fig:semtransfer}
\vskip -0.1in
\end{figure}

Following the shape transfer, our goal is to generate a texture using the extracted source and target UV texture flows $U_S$ and $U_T$ obtained from the encoder. However, birds often exhibit different textures or colors on different body parts, such as sparrows that have a white belly, a messy back pattern, and fixed head spots. Treating the overall texture as a single style during transfer may result in the transfer of local textures to incorrect semantic parts. To address this issue, we propose a semantic UV texture transfer method that performs semantic segmentation of the texture before transferring style to the corresponding semantic parts. 
Our approach involves creating a semantic texture using a semantic UV mask, which ensures consistency of semantic distributions as shown in Fig. \ref{fig:semtransfer}.  It consists of a semantic UV mask, a VGG encoder, a semantic stylizer, and a decoder.

\textbf{Semantic UV Mask} $U_{seg}\in \{1,2,3,4,5\}^{128\times256}$ is generated by SCOPS \cite{Hung2019scops}. To match the size of the subsequent VGG feature map, $U_{seg}$ is downsampled, but it is still referred to as $U_{seg}\in \mathbb{R}^{64\times128}$ for convenience. It comprises five parts, including four semantic UV parts: red head $U_{seg}^1\in \{0,1\}^{64\times128}$, green neck $U_{seg}^2\in \{0,1\}^{64\times128}$, blue belly $U_{seg}^3\in \{0,1\}^{64\times128}$ and yellow back $U_{seg}^4\in \{0,1\}^{64\times128}$, and one non-semantic part $U_{seg}^5\in \{0,1\}^{64\times128}$ to preserve the source UV texture. Each element $u_{jk}^i$ in $U_{seg}^i$ is defined as follows:  \begin{align}
u_{jk}^i=\left\{
\begin{aligned}
1&, & \text{if}\ u_{jk}=i,\\
0&, & \text{otherwise},
\end{aligned}
\right. 
\label{eq:maskelement}
\end{align}
where $1\leq i\leq 5$, $u_{jk}$ is the element of the UV mask $U_{seg}$.

\textbf{VGG Encoder} \cite{Simonyan2015vgg} is used to extract the feature of the UV texture, which has 512 channel maps of size $64\times128$. Given the source and target UV texture flows $U_S$ and $U_T$, the encoder outputs the features $V_S\in \mathbb{R}^{512\times64\times128}$ and $V_T\in \mathbb{R}^{512\times64\times128}$, respectively.



\textbf{Semantic Stylizer} is introduced to incorporate the semantic UV mask, represented as $U_{seg}$, into existing UV texture transformation methods for enhanced stylization. Specifically, we focus on the instance normalization approach, which aligns the channel-wise mean and variance of the masked source feature $V_S$ with those of the corresponding masked target feature $V_T$, without requiring additional parameters. We extend this approach to various methods such as AdaIN \cite{Huang2017adain}, linear style transfer (LST) \cite{Li2019lineartransfer} and EFDM \cite{zhang2022efdm}.
To handle the five semantic parts, \textit{semantic AdaIN} (SAdaIN) is computed by uniting these corresponding AdaIN feature matrix indexed by $U_{seg}$, that is, $\text{SA}\left(V_{S}, V_{T}, U_{seg}\right)=\sum_{i=1}^5\text{SA}^i\left(V_{S}^i, V_{T}^i, U_{seg}^i\right)$: 
\begin{align}
\text{SA}^i =\left\{
\begin{aligned}
\sigma(V_{T}^i) \frac{V_{S}^i\ominus\mu(V_{S}^i)}{\sigma(V_{S}^i)}+\mu(V_{T}^i)&, & \text{if}\ i\neq 5,\\
V_{S}^5&, & \text{if}\ i=5,
\end{aligned}
\right. 
\label{eq:madain}
\end{align}
where $V_{S}^i=V_{S}\odot \overline{U}_{seg}^i$, $V_{T}^i=V_{T}\odot \overline{U}_{seg}^i$,  $\overline{U}_{seg}^i$ is a binary matrix to repeat the index matrix $U_{seg}^i$ 512 times and reshape it to $512\times 64\times 128$, $\odot$ is an element-wise multiplication, $\ominus$, $\sigma$ and $\mu$ are the subtraction, variance, and mean computations only depend on the index binary matrix $\overline{U}_{seg}^i$. Here, $\text{SA}^5=V_{S}^5$ that means SAdaIN holds the source texture for the non-semantic part. Clearly, these local statistics are respectively computed across spatially masked locations, are merged into a new feature $\text{SA}$.

Similar to SAdaIN, \textit{semantic LST} (SLST) is defined as $\text{SL}\left(V_{S}, V_{T}, U_{seg}\right)=\sum_{i=1}^5\text{SL}^i\left(V_{S}^i, V_{T}^i, U_{seg}^i\right)$:
\begin{align}
\text{SL}^i =\left\{
\begin{aligned}
\left(\text{cov}(V_{S}^i)\text{cov}(V_{T}^i)V_{S}\right)\odot \overline{U}_{seg}^i&, & \text{if}\ i\neq 5,\\
V_{S}^5&, & \text{if}\ i=5,
\end{aligned}
\right. 
\label{eq:mlt}
\end{align}
where $\text{cov}(V_{S}^i)=V_{S}^i(V_{S}^i)^{\text{T}}$ is the centered covariance of $V_{S}^i$. 
 
In addition, we further extend to a histogram-based style transfer method, \ie, Exact Histogram Matching (EHM) \cite{coltuc2006exact}. Similarly, \textit{semantic EFDM} (SEFDM) is defined as
$\text{SE}\left(V_{S}, V_{T}, U_{seg}\right)=\sum_{i=1}^5\text{SE}^i\left(V_{S}^i, V_{T}^i, U_{seg}^i\right)$:
\begin{align}
\text{SL}^i =\left\{
\begin{aligned}
\text{EHM}\left(V_{S}^i,V_{T}^i\right)\odot \overline{U}_{seg}^i&, & \text{if}\ i\neq 5,\\
V_{S}^5&, & \text{if}\ i=5,
\end{aligned}
\right. 
\label{eq:mehm}
\end{align}

To increase the diversity of the texture, we also design a switch gate (SG) to stochastically exchange the part source feature $V_{S}^i$ and the part target feature $V_{T}^i$ in both the Eqs. \eqref{eq:madain} and \eqref{eq:mlt}. 

\textbf{Decoder} is identical to the decoder of AdaIN \cite{Huang2017adain} , LST \cite{Li2019lineartransfer} and EFDM \cite{zhang2022efdm} for our SAdaIN, SLST and SEFDM, respectively.


\subsection{Training Loss}
Our training loss is divided into two parts: shape transfer and texture transfer. The former mainly includes single-view 3D shape reconstruction with mask and perceptual loss and shape transfer with 3D key points loss. The latter part consists of traditional style transfer with style and content loss. These two parts are trained separately. The shared encoder and VGG-19 are pre-trained using UMR \cite{li2020self} and \cite{Simonyan2015vgg}, respectively. Given the source and target bird images $S$ and $T$, the shared encoder outputs the shape features $F_S$, $F_T$, and the UV maps $U_S$, $U_T$ respectively.

\textbf{Shape Transfer.} During source shape reconstruction, the model uses only the source shape feature $F_S$. However, our model requires both shape features ($F_S$ and $F_T$). Thus, we set both shape features to $F_S$ during reconstruction, and it outputs the result $O_S$. Similarly, they are set to $F_T$ for the target reconstruction and it outputs $O_T$. Following UMR \cite{li2020self}, the mask and perceptual losses are introduced as follows.

\textit{Mask Loss} aims to compute the negative IoU \cite{kato2019learning} between the ground truth instance mask $M$ and the predictive mask $\overline{M}$, which is produced by using \cite{liu2019soft} to render $O$ based on the camera pose predicted by the shared encoder \cite{li2020self}. Considering the source and target masks, it can be defined as: 
\begin{align}
\mathcal{L}_{mask}=\mathcal{L}_{m}(M_S,\overline{M}_S)+\mathcal{L}_{m}(M_T,\overline{M}_T), 
\end{align}
where $\mathcal{L}_{m}(M,\overline{M})=1-\frac{\|M \odot \overline{M}\|_{1}}{\|M+\overline{M}-M \odot \overline{M}\|_{1}}$. 

\textit{Perceptual Loss} calculates the perceptual distance \cite{zhang2018unreasonable} between the input image $I$ and the predictive image $\overline{I}$ produced by its UV map using the method \cite{li2020self}. Considering the source and target images $S$ and $T$, it is described as:
\begin{align}
\mathcal{L}_{per}=\mathcal{L}_{p}(S,\overline{S})+\mathcal{L}_{p}(T,\overline{T}),
\end{align}
where $\mathcal{L}_{p}(I,\overline{I})=\|\text{VGG}(I)-\text{VGG}(\overline{I})\|_{2}$. This loss can improve the visual quality of texture by capturing its details.


\textit{3D Keypoints Loss} is, more importantly, proposed to implement the shape transfer between source and target. The 3D key points are obtained by detecting the 2D key points \cite{guo2019aligned} from an image, determining their corresponding vertices on the reconstructed 3D shape using the predicted UV texture, and projecting them onto the symmetry plane. 
Since the symmetry plane is identical for all output 3D models, we can calculate the distance between key points based on their projections. 
This loss is defined as the $\ell_2$ distances of $\{p^{i},p_{S}^{i}, p_{T}^{i}\}_{i=1}^N$, which are the 3D keypoints of the transferred, source, and target shapes. It is calculated by 
\begin{align}
\mathcal{L}_{key}=\frac{1}{2 N} \sum_{i=1}^{N}\left(\|p^{i}-p_{S}^{i}\|_{2}+\lambda\|p^{i}-p_{T}^{i}\|_{2}\right),
\end{align}
where $N$ is the number of 3D keypoints, $\lambda$ is a balance parameter, and in this paper we set $N=15$ and $\lambda=1$. In summary, the overall objective for shape transformation is 
\begin{align}
\mathcal{L}_{shape}= \mathcal{L}_{mask}+\mathcal{L}_{per}+\mathcal{L}_{key}.
\end{align}

\textbf{Texture Transfer.} We extend AdaIN \cite{Huang2017adain}, LST \cite{Li2019lineartransfer}, and EFDM \cite{zhang2022efdm} into SAdaIN, SLST, and SEFDM respectively on the UV maps for the texture transfer. We utilize their style loss $\mathcal{L}_{style}$ and content loss $\mathcal{L}_{content}$ as our texture transfer loss $\mathcal{L}_{texture}=\mathcal{L}_{style}+\mathcal{L}_{content}$ on the UV textures $U_S$ and $U_T$. 



\begin{figure*}[t]
\centering
\includegraphics[width=0.87\linewidth]{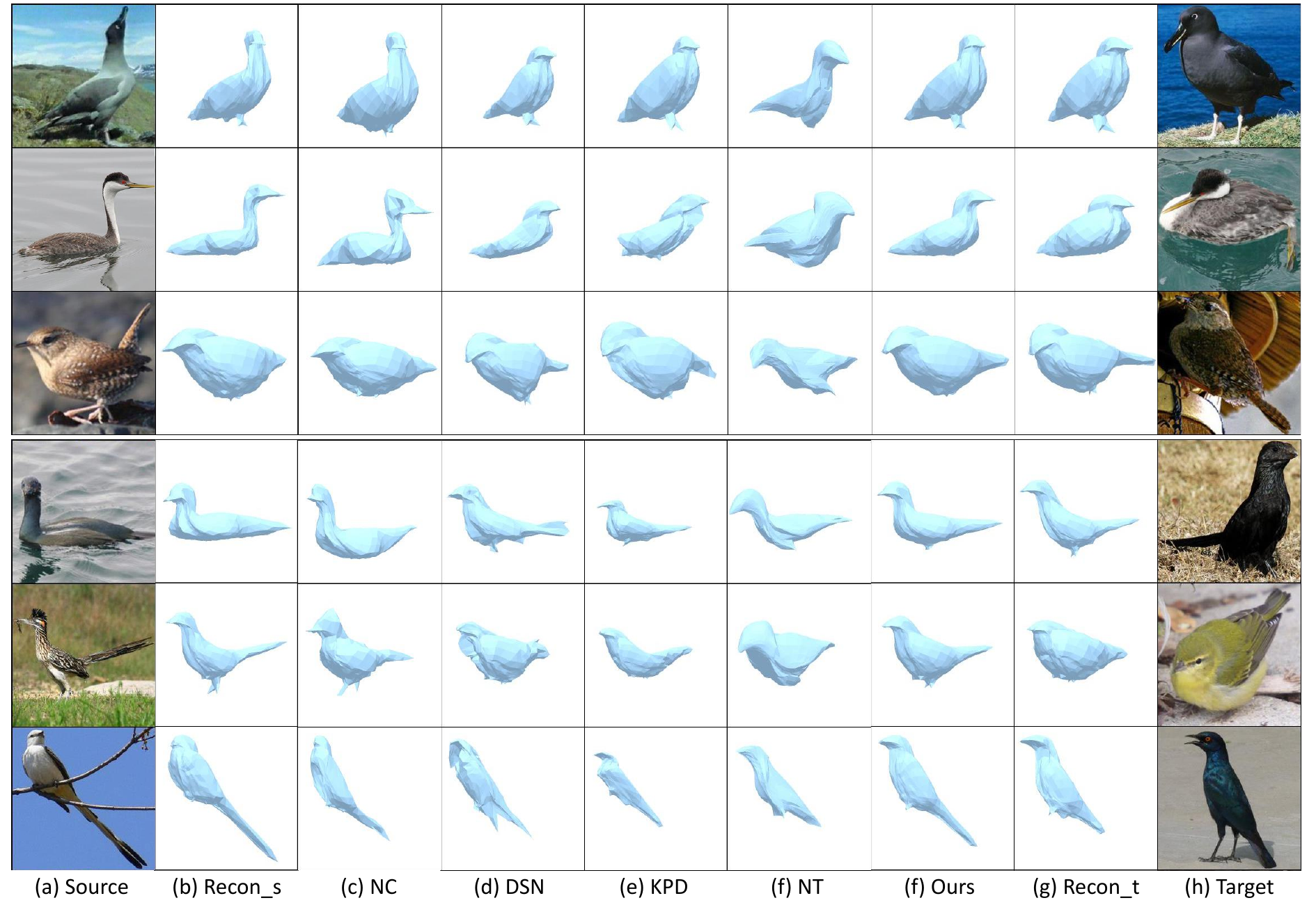}
\vskip -0.1in
\caption{Visual comparisons on shapes using 3D deformation methods (\eg, NC \cite{yifan2020neural}, DSN \cite{Sung2020DeformSyncNet}, KPD \cite{jakab2021keypointdeformer} and NT \cite{hui2022neural}). Both the source and target birds in the row 1-3 are from the same species, while they become from different species in the row 4-6. "Recon\_s" and "Recon\_t" represent the reconstruction of the source and the target respectively. Our method effectively achieves novel shape creation from these reconstructions and maintains reasonability. 
}
\label{fig:bird_shape}
\vskip -0.1in
\end{figure*}

\section{Experiments}
In this section, we conduct extensive experiments to evaluate the transfer ability of birds across populations of the same species and between species. \textit{More results are provided in supplementary materials.}

\textbf{Datasets.} Our method is validated on the CUB-200-2011 dataset \cite{WahCUB200}.
The dataset provides the bounding boxes and instance masks for each bird, and we obtain semantic segmentation using SCOPS \cite{Hung2019scops}. Based on the provided bounding boxes, each image is cropped and resized to 256x256 to form one training sample and randomly flipped with a probability of 0.5.

\textbf{Training.}
Our training schedule is divided into two steps due to the inclusion of texture transfer and shape transfer modules in our model. We first train the shape transfer network using the source and target images. Then, we use the UV textures produced by the encoder for the source and target to train the texture transfer network. In our experiment, these two steps are trained for 100k/80k iterations with an 8/8 batch size. We use Adam optimizer with a learning rate of 1e-4/initial 1e-4 and decay by 0.9995 per iteration. On a single GTX 2080ti GPU, training takes about 3/30 hours.

\subsection{Main Results}


To show the transfer capacity of our method for 3D bird creation, we compare with four shape deformation methods (\ie, NC \cite{yifan2020neural}, DSN \cite{Sung2020DeformSyncNet}, KPD \cite{jakab2021keypointdeformer} and NT \cite{hui2022neural}) and three texture style transformation methods (\ie, AdaIN \cite{Huang2017adain}, LST \cite{Li2019lineartransfer} and EFDM \cite{zhang2022efdm}).

\textbf{Visual comparisons.}
We validate the effectiveness of our model from two aspects: shape and texture transfer.


\textit{Shape Transfer.} Our work is not directly comparable to the 3D style transfer methods, even those involving 3D portrait stylization \cite{han2021exemplarbased} as they focus only on the face. For potential comparisons, we initially use the UMR to reconstruct the 3D models and then compare them with the shape deformation methods (\eg, NC, DSN, KPD, and NT).  
Fig. \ref{fig:bird_shape} showcases the results of the shape transformation. We observe that our method achieves a reasonable shape transformation, better matching the characteristics of the source and target. It demonstrates that when the shape difference between the source and target objects from different species is large, comparison methods can produce unreasonable distortions. This is likely due to the semantic part being difficult to align. Conversely, our method learns to reconstruct the 3D model to prevent shape distortion and transfer their shapes for novel shape generation simultaneously.

\begin{figure*}[t]
\centering
\includegraphics[width=0.90\linewidth]{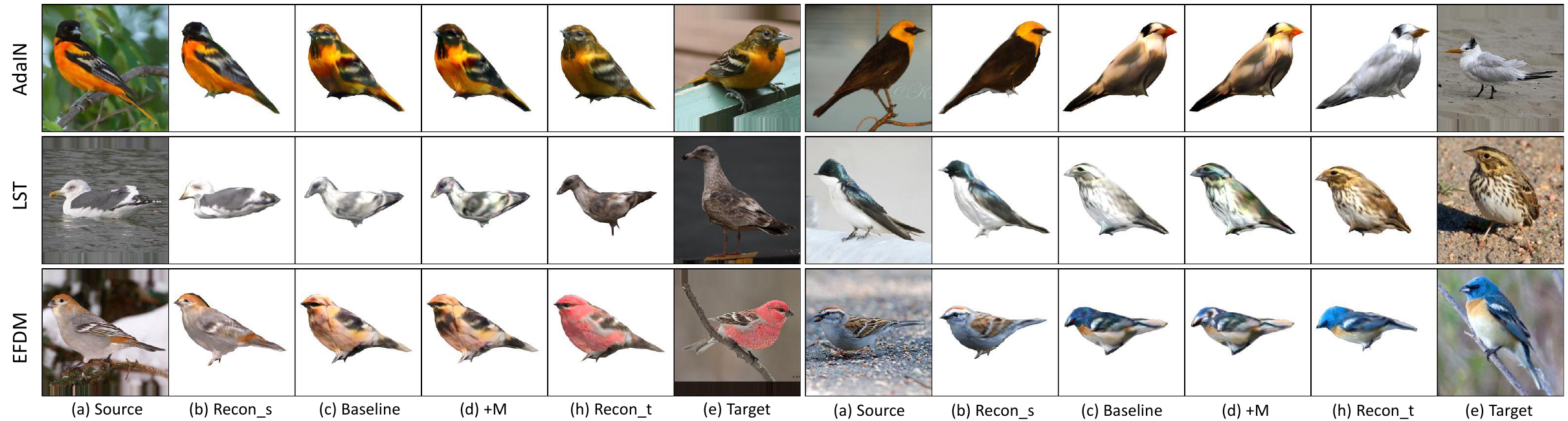}
\vskip -0.1in
\caption{Visual comparisons on textures using style transfer methods (\eg, AdaIN \cite{Huang2017adain}, LST \cite{Li2019lineartransfer} and EFDM \cite{zhang2022efdm}) associate with our method (\eg, semantic UV mask (+M)). Both the source and target birds in the left picture are from the same species, while they become from the different species in the right picture. We can see that a semantic UV mask can increase the diversity of the texture compared to the original methods. More results are shown in Fig. \ref{fig:switch_gate}.}
\label{fig:bird_tex}
\vskip -0.05in
\end{figure*}

\begin{figure}[t]
\centering
\includegraphics[width=0.9\linewidth]{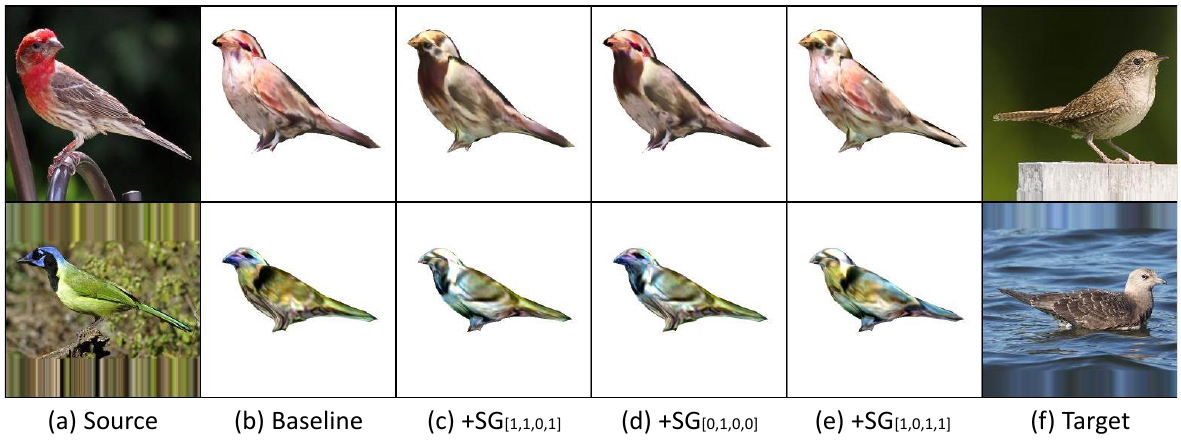}
\vskip -0.1in
\caption{Visual comparisons with different switch gates (+SG), the subscript are the switch signals for each semantic part (i.e. head, neck, belly, and back). It is obvious that the switch gate further increases the diversity of the transformed texture.
}
\label{fig:switch_gate}
\vskip -0.05in
\end{figure}

\textit{Texture Transfer.}
Fig. \ref{fig:bird_tex} shows the results of the style transfer algorithms (\eg, AdaIN, LST, and EFDM), after utilizing the modules of semantic transfer. Fig. \ref{fig:switch_gate} shows the result after adding different switch gates. Comparing Fig. \ref{fig:bird_tex} columns (c) and (d), we see that semantic transfer improves the effect of style transfer on each semantic part for all algorithms because the semantic mask prevents influence between different semantic parts. Furthermore, in Fig. \ref{fig:switch_gate} switch gate further enhances the diversity of results based on the semantic mask, making our transfer more consistent with natural evolutionary laws.


\begin{figure}[t]
	\centering
		\includegraphics[width=0.9\linewidth]{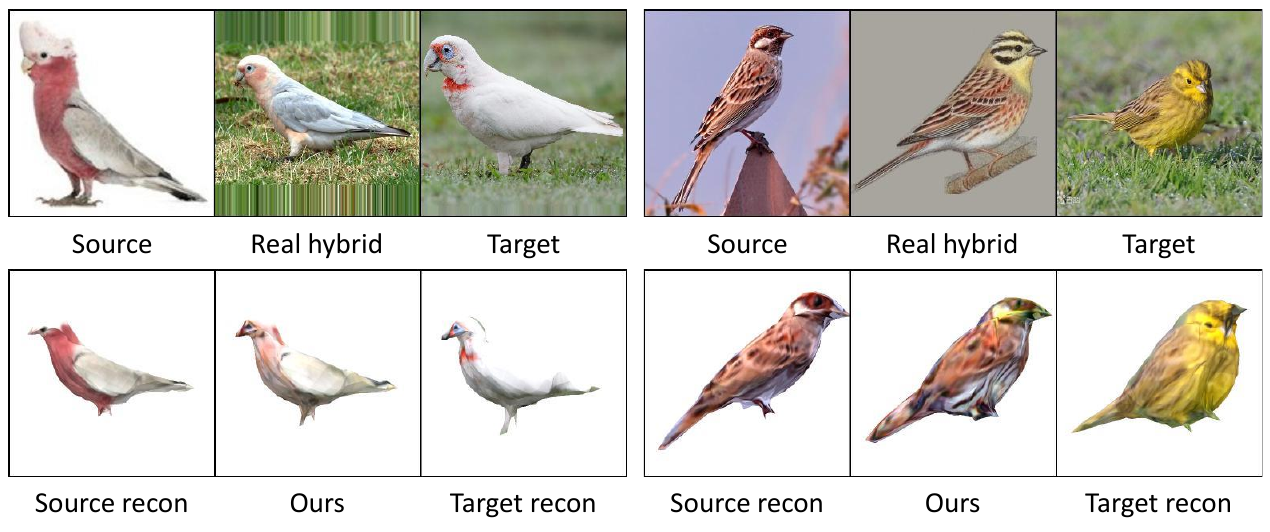}
		\vskip -0.1in
		\caption{Visual comparisons to the real hybrid birds.}\label{fig:horse}
		\vskip -0.15in
\end{figure}

\textit{Comparison with real hybrid birds.}
To further validate the effectiveness of our model, we simulate the morphological evolution of birds within and across species for biological research. We gather real hybrid birds and their parent species from the Internet \cite{buntingsandsparrows,hybrid_bird1}. Our model simulates hybridization using the parent species as the source and target input. As shown in Fig. \ref{fig:horse}, the synthesized results are very similar to the real hybrid examples. It's important to note that while our results are still far from the ultimate goal of biologists, this is a meaningful attempt to extend the style transfer into animal morphological evolution. 


\begin{table}[t]
  \centering
  \setlength{\tabcolsep}{2pt}
\renewcommand{\arraystretch}{1.1}
\resizebox{0.97\linewidth}{!}{
    \begin{tabular}{c|c|c|c|c|c}
    \Xhline{1.2pt} 
    Method & NC\cite{yifan2020neural} & DSN\cite{Sung2020DeformSyncNet} & KPD\cite{jakab2021keypointdeformer} & NT\cite{hui2022neural} & Ours \\
    \cline{1-6}
    Mask IoU$\uparrow$ & 0.6670 & 0.6937 & 0.5699 & 0.5209 & {\bf0.7316} \\
    \Xhline{1.2pt} 
    \end{tabular}}%
    \vskip -0.1in
      \caption{Quantitative evaluations of shape transfer methods.}
      \vskip -0.15in
  \label{tab:quant}%
\end{table}%

\textbf{Qualitative Results.}
We use mask IoU to assess shape transformation quality, comparing the Soft Rasterizer \cite{liu2019soft} output to ground truth masks of source and target. Our results outperform compared methods in preserving shape outline of both source and target birds (Table \ref{tab:quant}). Mask IoU within $0.7$ to $0.95$ is reasonable, as smaller values indicate loss of shape information, while higher values suggest lack of diversity. In addition, our goal with texture transformation is to improve the textural diversity of species, and we have yet to find a suitable metric to measure this.
\begin{table*}[tp]
  \centering
  \setlength{\tabcolsep}{2pt}
    \renewcommand{\arraystretch}{1.1}
    \resizebox{.9\linewidth}{!}{
    
    \begin{tabular}{c|cc|cc|cc|ccccc|cc}
    \Xhline{1.2pt} 
    \multirow{2}[1]{*}{Method} & \multicolumn{6}{c|}{Texture} &\multicolumn{5}{c|}{Shape} & \multicolumn{2}{c}{Overall} \\
    \cline{2-14}
& AdaIN\cite{Huang2017adain}   & SAdaIN   & LT\cite{Li2019lineartransfer}  & SLT & EFDM\cite{zhang2022efdm} & SEFDM & NC\cite{yifan2020neural} & DSN\cite{Sung2020DeformSyncNet}   & KPD\cite{jakab2021keypointdeformer}  & NT\cite{hui2022neural} & Ours   & Fake  & Real \\
    \cline{1-14}
    Votes$\uparrow$ & 118    & \textbf{392}    & 147 & \textbf{363}   & 183 & \textbf{327} & 142    & 60    & 18 & 22   & \textbf{268} & 134   & \textbf{376} \\
    \Xhline{1.2pt}     
    \end{tabular}
    }%
    \vskip -0.1in
    \caption{Quantitative evaluations of user study.}
    \vskip -0.05in
  \label{tab:userstudy1}%
\end{table*}%

\textbf{User Study}
We conducted a user study to better assess our model's performance compared to existing models. This study consisted of three main components: shape transformation, texture transformation, and realism judgment, which included 25 questions in total. As a result, we collected 102 questionnaire responses with a total of 2550 votes. Table \ref{tab:userstudy1} shows the number of votes received for each option. In the shape transformation comparison, 52.5$\%$ of users prefer our results compared to 27.8$\%$ for NC, 11.8$\%$ for DSN, 3.5$\%$ for KPD, and 4.3$\%$ for NT. In the texture transformation comparison, 76.9$\%$ of users prefer the results of SAdaIN over AdaIN, 71.2$\%$ prefer the results of SLST results over LST, and 64.1$\%$ prefer the results of SEFDM results over EFDM. In the realism judgment, 73.7$\%$ of users think that our result is more realistic.

\begin{figure*}[t]
\centering
\includegraphics[width=.73\linewidth]{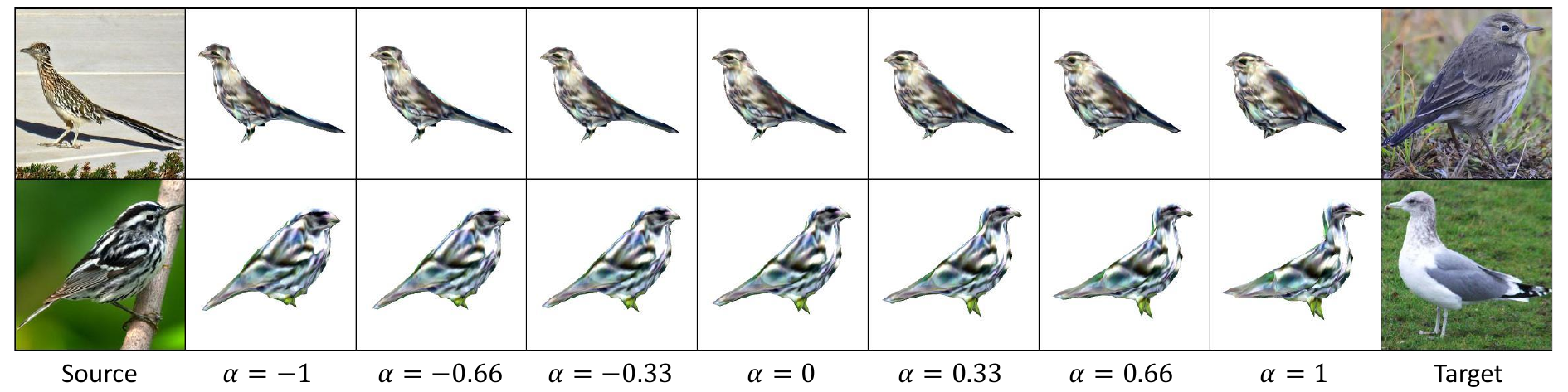}
\vskip -0.1in
\caption{Transfer degree of the shape controlled by the scale factor $\alpha$ from $-1$ to $1$.}
\label{fig:alpha}
\vskip -0.1in
\end{figure*}

\begin{figure}[t]
	\centering
		\includegraphics[width=0.9\linewidth]{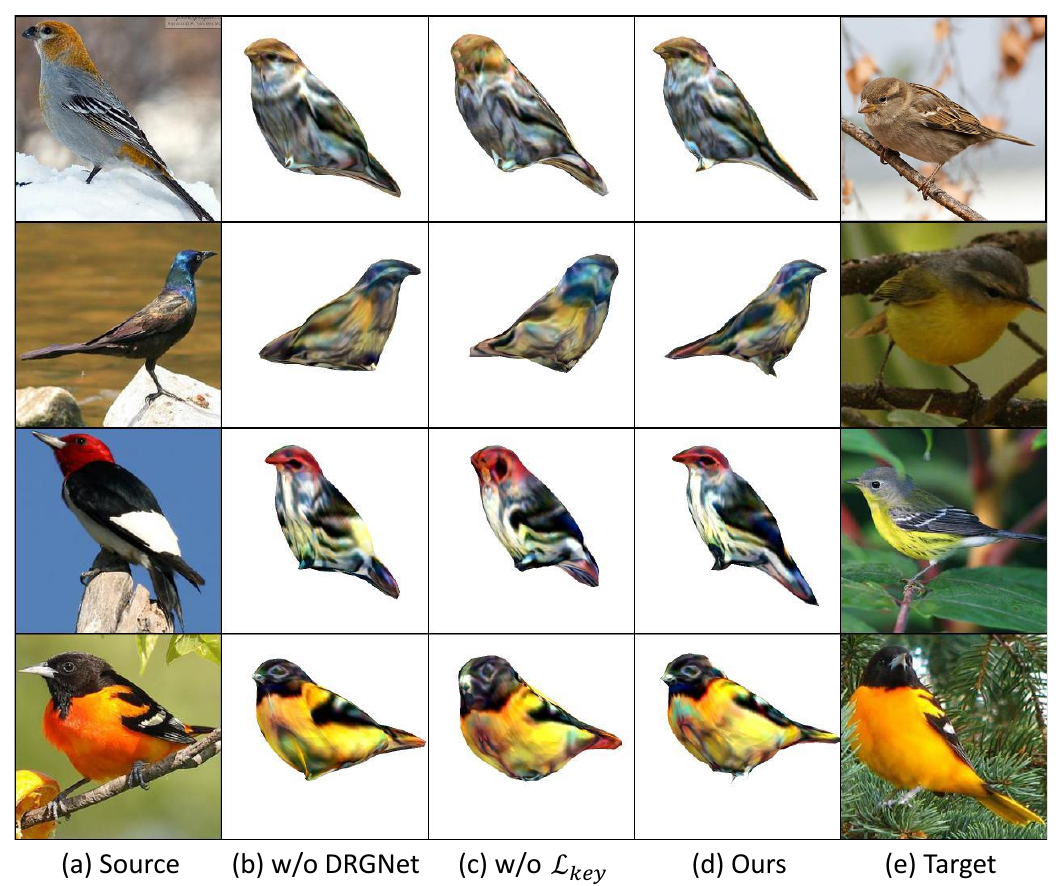}
		\vskip -0.1in
		\caption{Shape ablation study of DRGNet and 3D keypoints loss $\mathcal{L}_{key}$ using our SLST+SG as texture transformation method.}\label{fig:ablation}
		\vskip -0.1in
\end{figure}

\begin{figure}[t]
	\centering
		\includegraphics[width=0.9\linewidth]{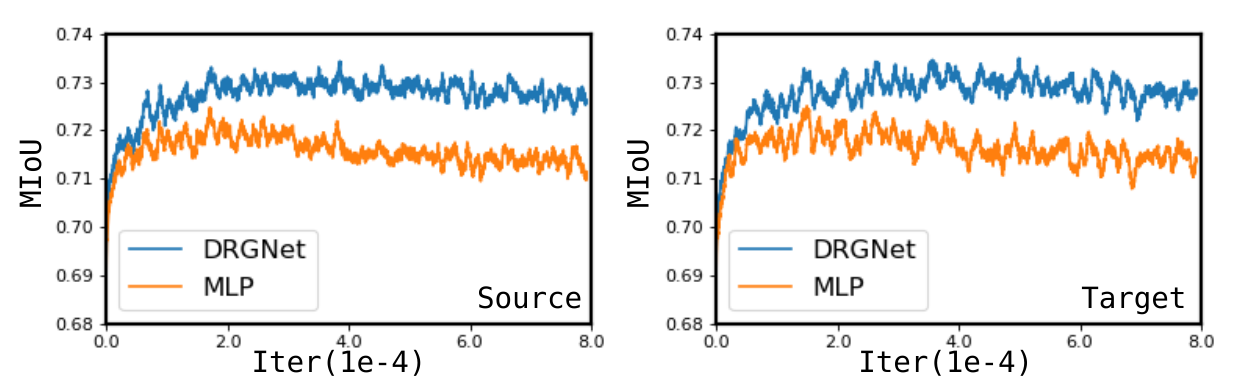}
		\vskip -0.1in
		\caption{Masked IoU using DRGNet and MLP.}\label{fig:miou}
		\vskip -0.1in
\end{figure}

\subsection{Ablation Study}
Since the modules of semantic transfer and switch gate have been compared in the previous subsections. Here we conduct ablation experiments on DRGNet, 3D keypoints loss $\mathcal{L}_{key}$ and scale factor $\alpha$, respectively.

\textbf{Effect of DRGNet.} We evaluate the importance of DRGNet by replacing it with MLP. From Fig. \ref{fig:ablation}, (e) ours using DRGNet can restore more edge details than (c) $w/o$ DRGNet (that is MLP) because it helps in the coordination of shape features. Moreover, Fig. \ref{fig:miou} also shows that DRGNet has a higher masked IoU than the MLP. 

\textbf{Effect of 3D keypoints loss $\mathcal{L}_{key}$.} We evaluate the effectiveness of $\mathcal{L}_{key}$ by replacing it with masked IoU loss, where the masks of the source and the target are used to constrain the result together. Then the results using the mask IoU loss instead of $\mathcal{L}_{key}$ are shown in Fig.\ref{fig:ablation} (d) $w/o$ $\mathcal{L}_{key}$, indicating significantly poorer outputs. The reason is that when the shape difference between the source and target object is large, their masks are more likely to have shape conflicts, resulting in an averaged result.

\textbf{Scale factor $\alpha$.} We use $\alpha$ to adjust the ratio between source and target features from $-1$ to $1$. Fig.\ref{fig:alpha} demonstrates that the scale parameter $\alpha$ effectively controls the presentation of the source and target characteristics in the results. When $\alpha=-1$ or $1$, the results are exactly the source or target 3D reconstruction. When $\alpha=0$, the results are to combine the half source with the half target. In addition, we also conducted experiments on DRGNet with different layers, which are included in the supplementary materials. 



\begin{figure}[t]
\centering
\includegraphics[width=0.9\linewidth]{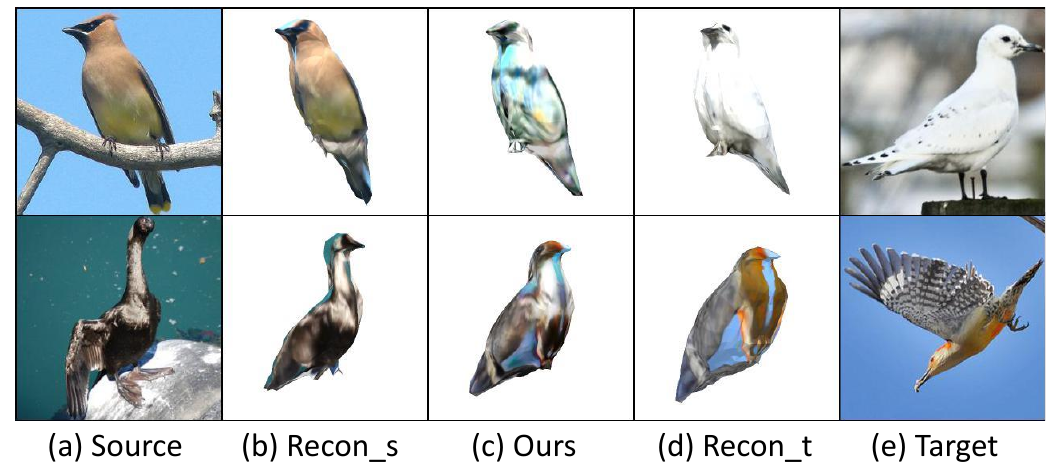}
\vskip -0.1in
\caption{Two examples of failed cases.}
\label{fig:limitation}
\vskip -0.1in
\end{figure}
\setlength{\intextsep}{0.25\baselineskip}
\textbf{Limitation.}
In this section, we discuss the limitations of our model. As we intend to perform style transformation on 2D images at a 3D level, the model needs to understand the 3D structure of the input image correctly. The model will produce problematic results when it incorrectly perceives the three-dimensional structure of the input image. According to Fig. \ref{fig:limitation}, the model improperly treats the sky as part of the bird, leading to a blue appearance. In the second row, the incorrect reconstruction of the complex target directly fails our style transfer algorithm.

\section{Conclusion}
In this paper, we proposed a style transfer method to automatically create novel 3D birds from single-view images.
We obtain the transformed shape features by filtering and coordinating the image features progressively and then utilizing a semantic transfer module to enhance the expressiveness of semantic texture style and the diversity of evolution.
Experimental results on the CUB dataset demonstrate that our method achieves state-of-the-art performance. 
Note that this is the first meaningful try to extend the style transfer into the novel 3D bird creations from the images. 
\section{Acknowledgments}
This work was partially supported by the National Science Fund of China (Grant Nos. 62072242, 62206134). 
{\small
\bibliographystyle{ieee_fullname}
\bibliography{egbib}
}

\end{document}


\title{Creative Birds: Self-supervised Single-view 3D Style Transfer\\
(Supplementary
Material)}


\maketitle

\begin{abstract}
This supplementary material consists of the following four parts: additional info regarding the loss function (section \ref{1}), a description of our evaluation interface for the user study (section \ref{2}), an experiment on DRGNet with different numbers of layers (section \ref{3}) and more comparison results (section \ref{4}).
\end{abstract}

\vspace{-4mm} 

\section{Additional Info on 3D Reconstruction Loss}\label{1}
In shape transformation, single-view 3D reconstruction is performed. Due to space constraints, only mask loss and perceptual loss are presented in the main paper. However, as in UMR \cite{li2020self}, we also use some other losses to improve the visual effect of the reconstructed results. 

Mask loss alone is insufficient for shape reconstruction because it only provides information about a single viewpoint. Therefore, we use deformation loss \cite{li2020self} to enable the model properly incorporate the 3D prior from the mesh template. In addition, graph laplacian constraint \cite{liu2019soft,cmrKanazawa2018,liu2019soft} and edge regularization \cite{wang2018pixel2mesh} are also employed to help smooth the reconstruct shape. 
In order to improve the visual effect of texture, we employ distance transform loss \cite{cmrKanazawa2018} to encourage texture flow select pixel inside the instance mask, and use texture cycle loss \cite{li2020self} to further optimize the position of the selected pixel.
Moreover, we ensure the semantic consistency \cite{li2020self} by constraining the chamfer distance and the l2 distance of each semantic part and its center between the input image and the rendered image. Lastly, multiple camera hypothesis \cite{tulsiani2018multi} is employed to avoid local minima (we used eight camera hypothesis here).

\begin{figure}[th]
\centering
\includegraphics[width=0.80\linewidth]{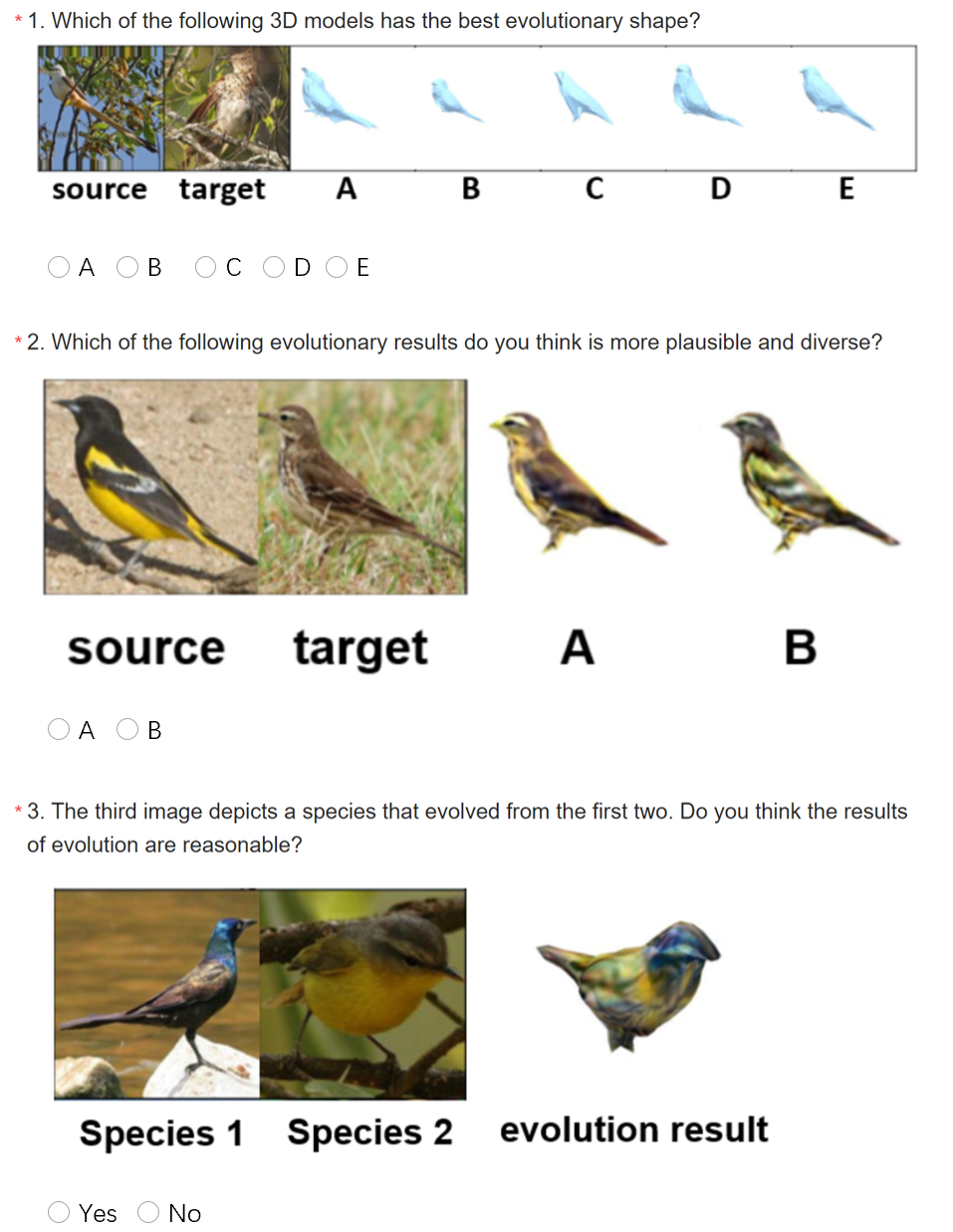}
\caption{Evaluation Interface.}
\label{fig:userstudy}
\vskip -0.1in
\end{figure}

\section{User Study}\label{2}
As stated in the "Experiments" section of our paper, we conducted a user study to compare our model to existing models. It consisted of three main components: a comparison of shape transformation (five questions about NC, DSN, KPD, NT and ours), a comparison of texture transformation (five questions about AdaIN and ours, five questions about LST and ours, five questions about EFDM and ours), and a judgement of realism (five T/F questions). Part of the evaluation interface is shown in Fig. \ref{fig:userstudy}.

\section{Additional Experiment on DRGNet}\label{3}
\begin{figure*}[th]
\centering
\includegraphics[width=0.97\linewidth]{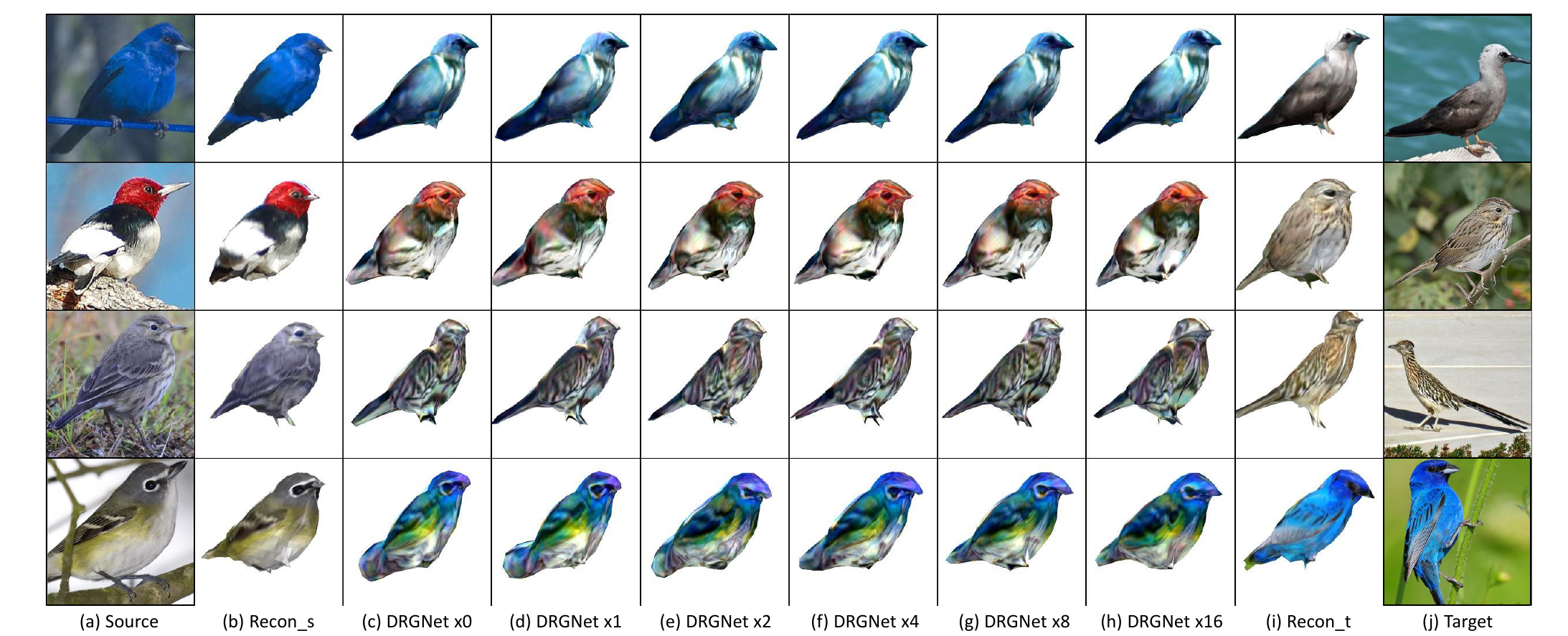}
\caption{Shape ablation study of DRGNet with different number of layers using our SLST+SG as texture transformation method.}
\label{fig:ablation}
\vskip -0.1in
\end{figure*}

After proving the efficacy of DRGNet for feature coordination, it remains unclear how many layers of DRGNet are needed for the transformation task, so here we conducted further experiments on the number of layers of DRGNet. The results are shown in Fig. \ref{fig:ablation}. 

\section{More Results}\label{4}
\begin{figure*}[h]
\centering
\includegraphics[width=\linewidth]{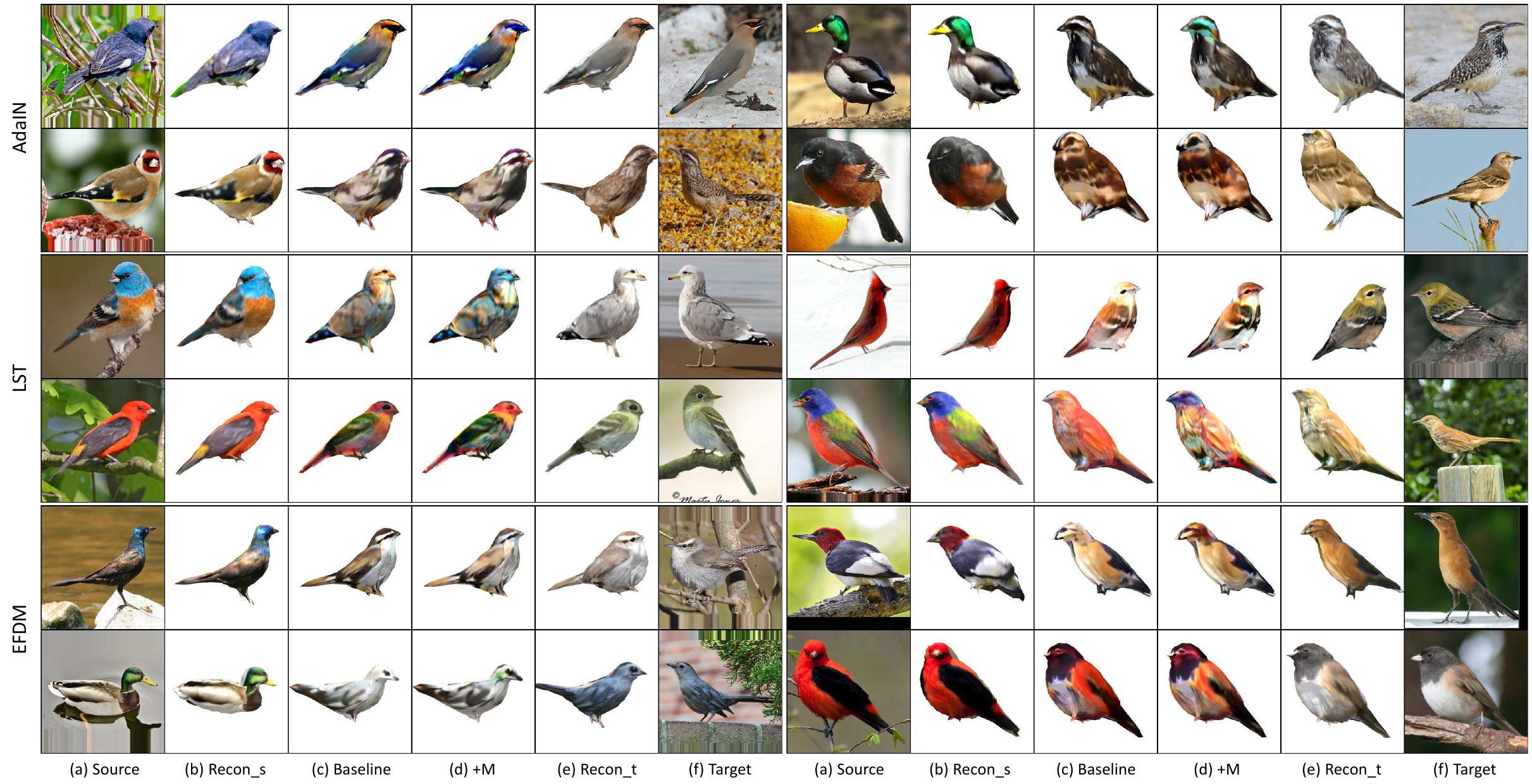}
\caption{More visual comparisons on textures using style transfer methods (\eg, AdaIN \cite{Huang2017adain}, LST \cite{Li2019lineartransfer}), EFDM \cite{zhang2022efdm} and our methods (\eg, semantic UV mask (+M)).}
\label{fig:tex}
\vskip -0.1in
\end{figure*}

\begin{figure*}[th]
\centering
\includegraphics[width=.90\linewidth]{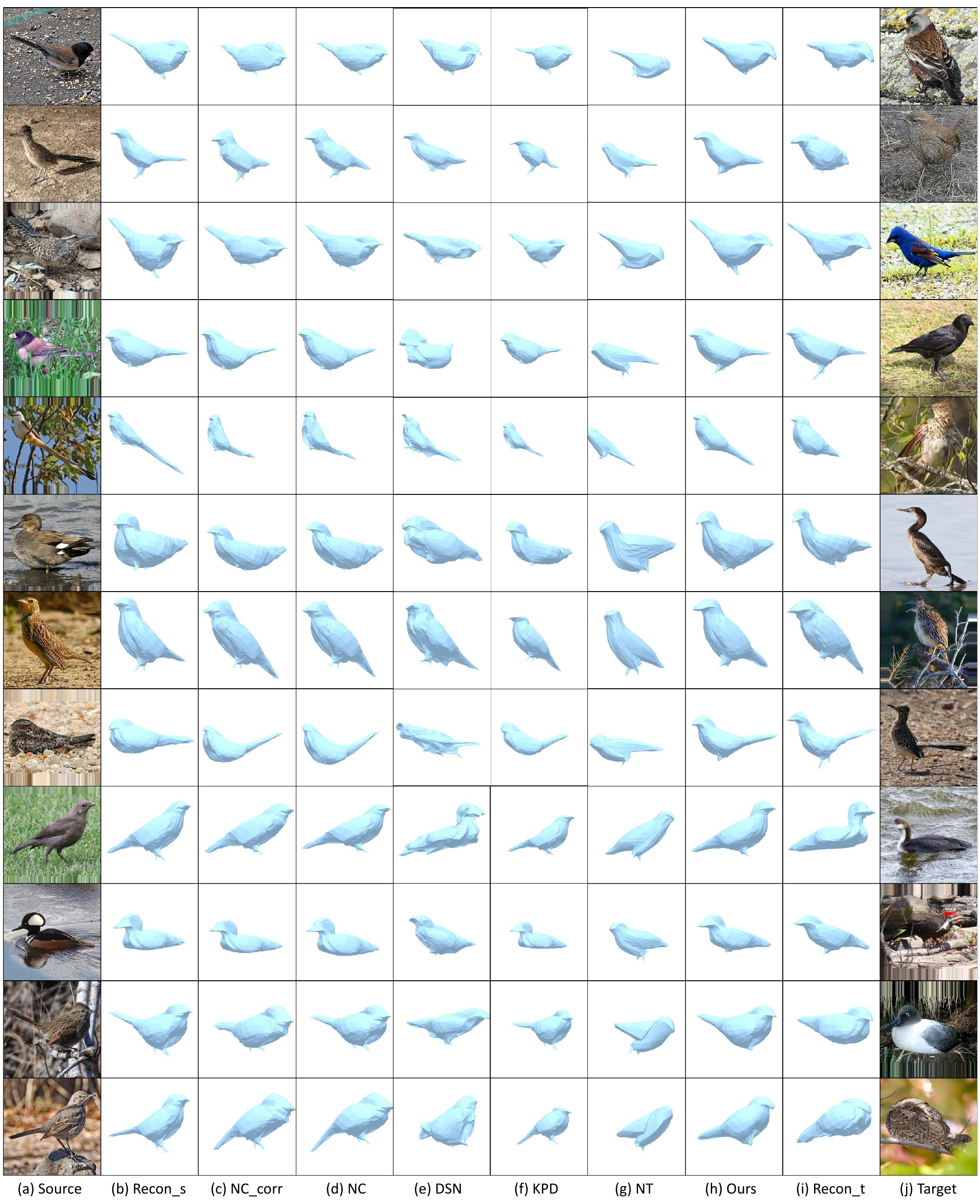}
\caption{More visual comparisons on shape using 3D shape deformation methods (\eg, NC \cite{yifan2020neural}, DSN \cite{Sung2020DeformSyncNet}, KPD \cite{jakab2021keypointdeformer}, NT \cite{hui2022neural} and our method.}
\label{fig:shape}
\vskip -0.1in
\end{figure*}

Here, we provide more results to help the reader evaluate the performance of our model. We show our comparison results from two aspects: 1. shape transformation (Fig. \ref{fig:shape}), 2. texture transformation (Fig. \ref{fig:tex}).   


\clearpage\clearpage

{\small
\bibliographystyle{ieee_fullname}
\bibliography{supplement}
}